\documentclass{article} % For LaTeX2e
\usepackage[preprint]{colm2026_conference}

\usepackage{microtype}
\usepackage{hyperref}
\usepackage{url}
\usepackage{booktabs}

\usepackage{url}

\usepackage{microtype}
\usepackage{url}
\usepackage{booktabs}
\usepackage{times}
\usepackage{latexsym}
\usepackage{amsmath}
\usepackage{amsfonts}
\usepackage{graphicx}
\usepackage{multirow}
\usepackage{subcaption}
\usepackage[T1]{fontenc}

\usepackage{tikz}

\usepackage{listings}
\usepackage{color}
\usepackage{minted}
\definecolor{dkgreen}{rgb}{0,0.6,0}
\definecolor{gray}{rgb}{0.5,0.5,0.5}
\definecolor{mauve}{rgb}{0.58,0,0.82}
\definecolor{lightblue}{rgb}{0.9, 0.95, 1.0}
\definecolor{lightgray}{rgb}{0.9, 0.9, 0.9}
\definecolor{mygreen}{RGB}{145,194,110}

\usepackage{enumitem}
\newcommand{\circled}[1]{\textcircled{\scriptsize{#1}}}
\usepackage{colortbl}
\usepackage{pifont}
\usepackage{algorithmic}
\usepackage{algorithm}

\usepackage[utf8]{inputenc}
\usepackage{multirow}
\usepackage{microtype}
\usepackage{url}

\usepackage{pifont}
\usepackage{microtype}
\usepackage{booktabs}
\usepackage{times}
\usepackage{latexsym}
\usepackage{amsmath}
\usepackage{amsfonts}
\usepackage{multirow}
\usepackage{microtype}
\usepackage{booktabs}
\usepackage{graphicx}
\usepackage{subcaption}
\usepackage[dvipsnames]{xcolor}
\usepackage{wrapfig}

\definecolor{purplecustom}{RGB}{105, 58, 123}
\definecolor{bluecustom}{RGB}{115, 127, 216}
\definecolor{goldcustom}{RGB}{234, 191, 121}
\definecolor{+}{rgb}{0.0, 0.6, 0.3} 
\definecolor{-}{rgb}{0.70,0.13,0.13}

\definecolor{darkblue}{rgb}{0, 0, 0.5}
\hypersetup{colorlinks=true, citecolor=darkblue, linkcolor=darkblue, urlcolor=darkblue}

\title{ImageEdit-R1: Boosting Multi-Agent Image Editing via Reinforcement Learning}

\author{Yiran Zhao$^{1}$ \quad Yaoqi Ye$^{1}$ \quad Xiang Liu$^{1}$ \quad Michael Qizhe Shieh$^{1}$\footnotemark[2] \quad Trung Bui$^{2}$\footnotemark[2] \\
  $^1$ National University of Singapore \quad $^2$ Adobe Research \\
}

\begin{document}

\ifcolmsubmission
\linenumbers
\fi

\maketitle
\renewcommand{\thefootnote}{\fnsymbol{footnote}}
\footnotetext[2]{Correspondence to: Michael Qizhe Shieh (\href{michaelshieh@comp.nus.edu.sg}{michaelshieh@comp.nus.edu.sg}) and Trung Bui (\href{bui@adobe.com}{bui@adobe.com}).}

\renewcommand{\thefootnote}{\arabic{footnote}}

\begin{abstract}
With the rapid advancement of commercial multi-modal models, image editing has garnered significant attention due to its widespread applicability in daily life. 
Despite impressive progress, existing image editing systems, particularly closed-source or proprietary models, often struggle with complex, indirect, or multi-step user instructions. These limitations hinder their ability to perform nuanced, context-aware edits that align with human intent.
In this work, we propose \texttt{ImageEdit-R1}, a multi-agent framework for intelligent image editing that leverages reinforcement learning to coordinate high-level decision-making across a set of specialized, pretrained vision-language and generative agents. Each agent is responsible for distinct capabilities—such as understanding user intent, identifying regions of interest, selecting appropriate editing actions, and synthesizing visual content—while reinforcement learning governs their collaboration to ensure coherent and goal-directed behavior. Unlike existing approaches that rely on monolithic models or hand-crafted pipelines, our method treats image editing as a sequential decision-making problem, enabling dynamic and context-aware editing strategies. Experimental results demonstrate that \texttt{ImageEdit-R1} consistently outperforms both individual closed-source diffusion models and alternative multi-agent framework baselines across multiple image editing datasets.\footnote[1]{Our code is publicly available at \url{https://github.com/zhaoyiran924/ImageEdit-R1}}
\end{abstract}

\section{Introduction}

Image editing~\citep{zhu2020domain, brooks2023instructpix2pix, kawar2023imagic} has long been a fundamental task in computer vision and graphics, playing a crucial role in applications such as creative design~\citep{dai2024implementation, nguyen2024epedit}, social media~\citep{lee2021social, agrawal2021impact}, e-commerce~\citep{dagan2023shop, ma2022ei}, and digital content production~\citep{wu2024research}. Especially, recent progress in large-scale vision-language models~\citep{hurst2024gpt, comanici2025gemini, li2025survey} (VLMs) and generative diffusion models~\citep{cao2024survey} has significantly advanced image manipulation capabilities.
However, despite these impressive strides, current systems—especially proprietary or closed-source models—often falter when faced with complex, compositional, or multi-step user instructions~\citep{sushko2025realedit, taesiri2025understanding, zhao2024ultraedit}. These limitations hinder their ability to produce context-aware edits that faithfully reflect user intent, particularly in real-world scenarios where instructions are often indirect or ambiguous.
On the other hand, the professional image editing software provides a comprehensive set of tools to edit images and preserve object identities. However, identifying the best workflow requires human expertise and manual process~\citep{bradley2024mastering, kelby2023photoshop, gosselin2022hand}. 
\begin{figure*}[t]
    \centering
    \includegraphics[width=0.98\textwidth]{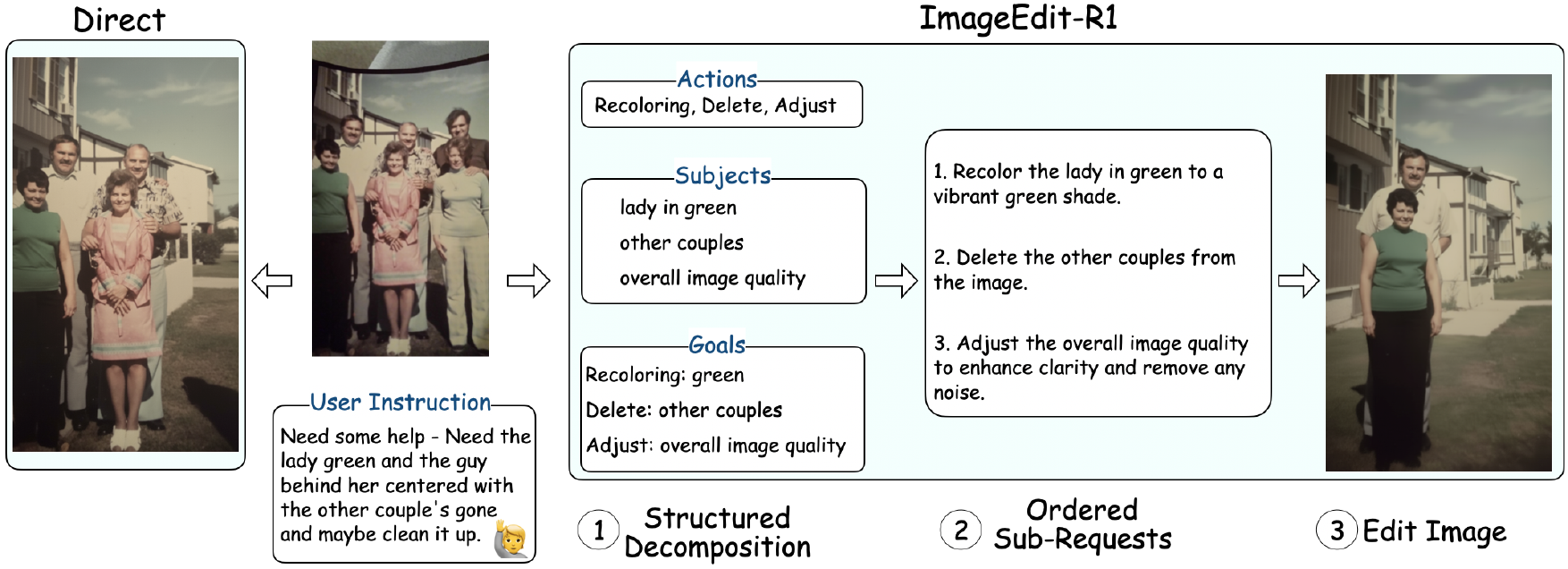}
    \caption{Overview of \texttt{ImageEdit-R1}: \circled{1} The \textit{decomposition agent} analyzes the user instruction and input image to extract a structured representation of the desired edits, including editing actions, subjects, and goals. \circled{2} The \textit{sequencing agent} arranges these components into an ordered list of sub-requests, enabling interpretable and modular execution. \circled{3} The \textit{editing agent}, built on a diffusion model, performs the actual image edits by sequentially applying the sub-requests. }
    \label{fig:framework}
\vspace{-0.2cm}
\end{figure*}

In this paper, we introduce a multi-agent framework for intelligent image editing, \texttt{ImageEdit-R1}, which formulates the editing process as a sequential decision-making problem and employs reinforcement learning to enhance agent collaboration. As shown in Figure~\ref{fig:framework}, \texttt{ImageEdit-R1} consists of three specialized agents that work together to interpret user intent, plan editing steps, and generate high-quality visual outputs.
The first component is the \textit{decomposition agent}, which analyzes both the user request and the image to extract a structured representation of the desired edits. This includes identifying the editing actions, relevant visual subjects, and the intended goals. To improve the accuracy and robustness of this decomposition, we apply Group Relative Policy Optimization (GRPO)~\citep{shao2024deepseekmath} with a set of carefully designed rewards that encourage correct formatting and semantic consistency with ground truth annotations.
Next, the \textit{sequencing agent} organizes the extracted edit components into an ordered sequence of sub-requests. This step enables interpretable and controllable execution by breaking down complex instructions into manageable tasks. Finally, the \textit{editing agent}, based on a diffusion model, performs the actual image modifications by following the generated sequence of sub-requests.
By coordinating these agents through reinforcement learning and structured reasoning, \texttt{ImageEdit-R1} supports compositional, interpretable, and context-aware image editing. 

% This approach moves beyond monolithic systems by enabling dynamic planning and fine-grained control, making it well-suited for real-world editing scenarios with ambiguous or multi-step user instructions.

We conduct extensive experiments to evaluate the effectiveness of \texttt{ImageEdit-R1} across a diverse set of image editing models and benchmark datasets. Specifically, we assess performance on three challenging multi-turn instruction editing datasets: PSR~\citep{taesiri2025understanding}, RealEdit~\citep{sushko2025realedit}, and UltraEdit~\citep{zhao2024ultraedit}, using both GPT-4o~\citep{hurst2024gpt} and Gemini-2.5~\citep{comanici2025gemini} as evaluators.  
% Our method consistently improves performance over strong baselines by leveraging a multi-agent framework with instruction decomposition and reinforcement learning. Notably, \texttt{ImageEdit-R1} achieves substantial gains across all backbones, including FLUX.1-Kontext-dev~\citep{labs2025flux1kontextflowmatching}, Qwen-Image-Edit~\citep{wu2025qwenimagetechnicalreport}, and NanoBanana~\citep{comanici2025gemini}. For example, on FLUX.1-Kontext-dev, it improves the average score from 7.21 to 8.23, a +1.02 increase. On Qwen-Image-Edit and NanoBanana, it achieves +0.46 and +0.34 improvements, respectively, over their non-RL counterparts. These gains are achieved without modifying the underlying image editing models, indicating that our framework generalizes well across architectures. The results demonstrate that \texttt{ImageEdit-R1} effectively enhances instruction alignment and visual quality, making it a strong and flexible solution for complex, multi-step image editing tasks.
Our method consistently improves over strong baselines by incorporating a multi-agent framework with instruction decomposition and reinforcement learning. Within the full evaluation range of 0 to 10, \texttt{ImageEdit-R1} achieves substantial gains across all editing backbones. For instance, on FLUX.1-Kontext-dev~\citep{labs2025flux1kontextflowmatching}, \texttt{ImageEdit-R1} improves the average score from 7.21 to 8.23, a gain of +1.02. Similarly, it boosts performance on Qwen-Image-Edit~\citep{wu2025qwenimagetechnicalreport} from 8.39 to 8.85 (+0.46), and on NanoBanana~\citep{comanici2025gemini} from 8.32 to 8.66 (+0.34). These gains are achieved without modifying the underlying image editing models, demonstrating that our approach generalizes effectively across architectures and tasks. Overall, the results highlight that \texttt{ImageEdit-R1} significantly enhances instruction alignment, visual quality, and content preservation for complex image editing.

\section{Method}

In this section, we present the multi-agent framework and explain how to enhance it using reinforcement learning.

\subsection{Overview}

To effectively handle complex user requests \( R \), we propose a structured decomposition approach that extracts three key components from each request: the \textit{edit actions}, \textit{edit subjects}, and the \textit{edit goals}. This structured representation facilitates precise understanding and controllable execution of image edits. In addition to the textual request, the original image \( I \) is incorporated to provide necessary visual context for accurate interpretation. To perform this decomposition, we employ a vision-language model, denoted as the decomposition agent \( \mathcal{A}_{\text{decom}} \), which takes \( R \) and \( I \) as its inputs and produces as its outputs a structured tuple:
\begin{equation}
  (\mathcal{R}_{\text{actions}}, \mathcal{R}_{\text{subjects}}, \mathcal{R}_{\text{goals}}) = \mathcal{A}_{\text{decom}}(R, I).
\end{equation}
For example, given the request \textit{``Change the color of her coat and possibly hair to scarlet or copper red''}, the decomposition yields action: \textit{[``Recoloring'']}, subjects: \textit{[``coat'', ``hair'']}, and goal: \textit{[``scarlet or copper red'']}. 
% This structured form enables downstream modules to interpret and execute user intent in a modular and interpretable manner.

Given the structured decomposition results, we further employ a sequencing agent \( \mathcal{A}_{\text{order}} \) to generate an ordered list of sub-requests based on the extracted actions, subjects, and goals. Formally, this process is defined as:
\begin{equation}
   \{r_1, \cdots, r_n\} = \mathcal{A}_{\text{order}}\big(I, (\mathcal{R}_{\text{actions}}, \mathcal{R}_{\text{subjects}}, \mathcal{R}_{\text{goals}})\big).
\end{equation}
Continuing the previous example, where the user request was \textit{``Change the color of her coat and possibly hair to scarlet or copper red''}, the agent generates a sequence of sub-requests such as: \textit{[``Recoloring coat to scarlet'', ``Recoloring hair to copper red'']}. 
% This ordering facilitates step-by-step execution and improves the interpretability and controllability of the editing process.

Finally, we employ a diffusion-based image editing model, referred to as the editing agent \( \mathcal{A}_{\text{edit}} \), to generate the modified image by sequentially executing the ordered sub-requests. Formally, the edited image \( I_{\text{new}} \) is obtained as:
\begin{equation}
  I_{\text{new}} = \mathcal{A}_{\text{edit}}\big(I, \{r_1, \cdots, r_n\}\big).
\end{equation}
% This step integrates the original image \( I \) and the structured instruction sequence to produce high-quality edits that align with the user’s intent, completing the multi-agent editing pipeline.

\subsection{Enhance through Reinforcement Learning}

To further improve the performance of the framework, we employ RL to enhance the capabilities of the agents, with a particular focus on the decomposition agent $\mathcal{A}_{\text{decom}}$.

\paragraph{Rewards} To encourage more accurate decomposition, we introduce four rewards corresponding to three decomposition components.
\begin{itemize}
    \item \textbf{Format Reward:} The model is incentivized to structure its output according to predefined formatting guidelines. Specifically, it must enclose the reasoning process within \texttt{<think>} and \texttt{</think>}, actions within \texttt{<action>} and \texttt{</action>}, subjects within \texttt{<subjects>} and \texttt{</subjects>}, and goals within \texttt{<goals>} and \texttt{</goals>}.

    \item \textbf{Action, Subject, and Goal Rewards:} Since these components are represented as sets, we evaluate the model's predictions using the F1-score rather than exact match, allowing for a more flexible and informative assessment of precision and recall against the ground truth.
\end{itemize}

\begin{figure}[t]
\centering
\scalebox{0.9}{
\begin{minipage}{\linewidth}
\begin{algorithm}[H]
\caption{\texttt{ImageEdit-R1}}
\renewcommand{\algorithmicrequire}{\textbf{Input:}}
\renewcommand{\algorithmicensure}{\textbf{Output:}}
\begin{algorithmic}[1]
  \REQUIRE Image-editing request $R$, input image $I$, decomposition agent $\mathcal{A}_{\text{decom}}$, sequencing agent $\mathcal{A}_{\text{order}}$, editing agent $\mathcal{A}_{\text{edit}}$, RL dataset $\mathcal{D}_{\text{RL}}$

  \STATE \texttt{// Reinforcement Learning for $\mathcal{A}_{\text{decom}}$}
  \FOR{$q \in \mathcal{D}_{\text{RL}}$}
    \STATE $\pi_{\theta_{\text{old}}} \leftarrow \pi_{\theta}$
    \STATE Sample $N$ trajectories $\{\tau_1, \cdots, \tau_N\} \sim \pi_{\theta_{\text{old}}}(q)$
    \STATE Compute rewards $R(\tau_i)$: format, action, subject, and goal rewards
    \STATE Compute normalized advantages $A_i$ and weights $r_{i,t}(\theta)$
    \STATE Update $\pi_{\theta}$ using GRPO loss $\mathcal{L}_{\text{GRPO}}$
  \ENDFOR
  
  \STATE \texttt{// Step 1: Structured Decomposition}
  \STATE $(\mathcal{R}_{\text{actions}}, \mathcal{R}_{\text{subjects}}, \mathcal{R}_{\text{goals}}) \leftarrow \mathcal{A}_{\text{decom}}(R, I)$

  \STATE \texttt{// Step 2: Generate Ordered Sub-Requests}
  \STATE $\{r_1, \cdots, r_n\} \leftarrow \mathcal{A}_{\text{order}}\big(I, (\mathcal{R}_{\text{actions}}, \mathcal{R}_{\text{subjects}}, \mathcal{R}_{\text{goals}})\big)$

  \STATE \texttt{// Step 3: Edit Image}
  \STATE $I_{\text{new}} \leftarrow \mathcal{A}_{\text{edit}}(I, \{r_1, \cdots, r_n\})$

  \ENSURE Edited image $I_{\text{new}}$
\end{algorithmic}
\label{alg:image_edit}
\end{algorithm}
\end{minipage}
}
\end{figure}

\paragraph{Reinforcement Learning} With the proposed reward design before, we adopt GRPO~\citep{shao2024deepseekmath} to train $\pi_{\theta}$  on the RL dataset $\mathcal{D}_{\text{RL}} = \{ q_1, q_2, \cdots, q_n \}$.
% \shafiq{let's put the elements.... Should we use $x_i$ or $c_i$ to be consistent (in stead of $q$?) }.
%With the proposed reward design, we primarily adopt GRPO~\cite{shao2024deepseekmath} to train the model $\mathcal{LLM}$ with patameters $\pi_{\theta}$ on the reinforcement learning dataset $\mathcal{D}_{\text{RL}}$. 
Specifically, for $q\in\mathcal{D}_{\text{RL}}$, we use the old policy from previous step $\pi_{\theta_{\text{old}}}$ to sample a group of $N$ individual responses $\tau_i$. Then, the RL loss is defined as:
\begin{equation}
\begin{aligned}
\mathcal{L}_{\text{GRPO}}(\theta) =\;& \mathbb{E}_{\tau_i \sim \pi_{\theta_{\text{old}}}(q),q\sim \mathcal{D_{\text{RL}}}} \frac{1}{\sum_{i=1}^N {\color{black}|\tau_i|}} \sum_{i=1}^N \sum_{t=1}^{|\tau_i|} \\
& \text{CLIP}(r_{i,t}(\theta), A_i, \epsilon) - \beta \cdot \mathbb{D}_{\text{KL}}[\pi_\theta \| \pi_{\text{ref}}]. \\%+ \gamma \cdot \mathcal{H}(\pi_{\theta}),  \\
% \text{where } {r_{i,t}(\theta) = \frac{\pi_\theta(\tau_{i,t} | q, \tau_{i,<t})}{\pi_{\theta_{\text{old}}}(\tau_{i,t} | q, \tau_{i,<t})}}.
\end{aligned}
\label{eq:grpo}
\end{equation}
where 
\begin{equation*}
    A_i = \frac{R(\tau_i) - \text{mean}(\{R(\tau_i) \mid \tau_i \sim \pi_{\theta_{\text{old}}}(\tau), i = 1,\ldots,N\})}{\color{black}\text{std}(\{R(\tau_i) \mid \tau_i \sim \pi_{\theta_{\text{old}}}(\tau), i = 1,\ldots,N\})},
\end{equation*}
and ${r_{i,t}(\theta) = {\pi_\theta(\tau_{i,t} | q, \tau_{i,<t})}/{\pi_{\theta_{\text{old}}}(\tau_{i,t} | q, \tau_{i,<t})}}$.

\subsection{{ImageEdit-R1}}

With the enhanced decomposition agent \( \mathcal{A}_{\text{decom}} \), we construct the \texttt{ImageEdit-R1} through collaborative multi-agent processing. Specifically, a user request \( R \) and its corresponding image \( I \) are first passed to \( \mathcal{A}_{\text{decom}} \), which decomposes the request into edit actions, subjects, and goals. These structured components are then fed into the sequencing agent \( \mathcal{A}_{\text{order}} \) to produce a coherent sequence of sub-requests. Finally, the editing agent \( \mathcal{A}_{\text{edit}} \) takes the original image \( I \) along with the generated sub-requests to produce the final edited image. Details are illustrated in Algorithm \ref{alg:image_edit}

\section{Experiment}

% \subsection{Setup}

% \paragraph{Models}

% \begin{itemize}
%     \item Decomposition Model: Qwen2.5-VL

%     \item Edit Model: FLUX.1-Kontext-dev, Qwen-edit, Nano-banana 

%     \item Evaluation Model: Gemini
    
% \end{itemize}

% \paragraph{Dataset}

% PSR, RealEdit (ablation on how RL can help), UltraEdit

% PSR + others for RL

% \paragraph{Baselines}

% \begin{itemize}
%     \item Original

%     \item Qwen3-VL

%     \item Open-source models

% \end{itemize}

\subsection{Experiment Setup}

\paragraph{Backbone Models} For both the decomposition and sequencing agents, we utilize Qwen2.5-VL-7B-Instruct~\citep{bai2025qwen25vltechnicalreport} as the unified backbone model across all experimental settings. For the image editing component, we evaluate three distinct backbone models: FLUX.1-Kontext-dev~\citep{labs2025flux1kontextflowmatching}, Qwen-Image-Edit~\citep{wu2025qwenimagetechnicalreport}, and Nano Banana. 
% \trung{"Check if Nano Bana is an open-source or not"}

\paragraph{Baselines} We evaluate our method against a range of state-of-the-art image editing baselines, including both open-source single-model approaches and proprietary systems. For single-model diffusion transformer (DiT) baselines, we consider Step1X-Edit~\citep{liu2025step1xeditpracticalframeworkgeneral}, ILLUME+~\citep{huang2025illumeilluminatingunifiedmllm}, and ICEdit~\citep{zhang2025incontexteditenablinginstructional}. For proprietary models, we compare against GPT-4o~\citep{hurst2024gpt} and SeedEdit~\citep{shi2024seededitalignimageregeneration}.
Additionally, within the context of multi-agent frameworks, we benchmark our approach against two variants: (1) a single model without any task-specific training (Original), and (2) the ImageEdit-R1 framework without reinforcement learning (ImageEdit-R1 (w/o RL)).

\paragraph{Benchmarks} To evaluate image editing proficiency, we utilize three established benchmarks: PSR~\citep{taesiri2025understandinggenerativeaicapabilities}, which includes 328 test-time editing requests spanning a wide range of difficulty levels; RealEdit~\citep{sushko2025realedit}, comprising approximately 9.3K test samples, from which we randomly sample 1,000 instances to ensure evaluation efficiency; and UltraEdit~\citep{zhao2024ultraedit}, which contains over 4 million edits, from which we likewise sample 1,000 examples for practical evaluation due to computational constraints.

\begin{table*}[t]
  \centering
\renewcommand{\arraystretch}{1.3}
\caption{Main results of \texttt{ImageEdit-R1} with evaluation range of 0 to 10. ``ImageEdit-R1 (w/ RL)'' refers to the variant \textit{without} RL for the decomposition agent. The ``Average'' column reports the mean score across all three benchmarks and two evaluators. Relative improvements over the original model are shown in superscript. \textbf{Bold} indicates the best performance, while \underline{underline} represent the second-best.}
\footnotesize
\setlength{\tabcolsep}{2.3pt}
\scalebox{0.88}{
  \begin{tabular}{c|l|cc|cc|cc|l}
    \toprule
   \multirow{2}{*}{\textbf{\normalsize{Edit Model}}} & \multicolumn{1}{c}{\multirow{2}{*}{\textbf{\normalsize{Method}}}} \vline &  \multicolumn{2}{c}{\textbf{\normalsize{PSR}}} \vline & \multicolumn{2}{c}{\textbf{\normalsize{RealEdit}}} \vline & \multicolumn{2}{c}{\textbf{\normalsize{UltraEdit}}} \vline & \multirow{2}{*}{\textbf{\normalsize{Average}}}  \\
    & & GPT-4o & Gemini-2.5 & GPT-4o & Gemini-2.5 & GPT-4o & Gemini-2.5  &   \\
    \midrule 
       \rowcolor{lightgray}\multicolumn{9}{c}{\textit{\normalsize{Single Model}}} \\
      \multirow{3}{*}{\textbf{\normalsize{\shortstack{Single\\-DiT-model}}}}  
      % & \normalsize{ReasonBrain~\cite{he2025reasoningedithypotheticalinstructionbased}} & & & & & & & \\ 
     % & \normalsize{Step1X-Edit~\cite{liu2025step1xeditpracticalframeworkgeneral}} & 6.70 & 6.58 & 6.83 & 6.92 & 7.51 & 7.67 & 7.04 \\ 
     % & \normalsize{ILLUME+~\cite{huang2025illumeilluminatingunifiedmllm}} & 6.00 & 6.22 & 6.37 & 6.50 & 7.28 & 6.89 & 6.54 \\
     % & \normalsize{ICEdit~\cite{zhang2025incontexteditenablinginstructional}} & 5.64 & 5.73 & 5.75 & 6.26 & 7.44 & 7.13 & 6.33 \\ \midrule
     & \normalsize{Step1X-Edit} & 6.70 & 6.58 & 6.83 & 6.92 & 7.51 & 7.67 & 7.04 \\ 
     & \normalsize{ILLUME+} & 6.00 & 6.22 & 6.37 & 6.50 & 7.28 & 6.89 & 6.54 \\
     & \normalsize{ICEdit} & 5.64 & 5.73 & 5.75 & 6.26 & 7.44 & 7.13 & 6.33 \\ \midrule
      \multirow{2}{*}{\textbf{\normalsize{\shortstack{Close\\-source-model}}}}  & \normalsize{GPT-4o} & \textbf{9.12} & 7.96 & 8.35 & 8.19 & 8.71 & 8.48 & 8.47 \\ 
     % & \normalsize{Gemini-2.5} & & & & & & & \\ 
     & \normalsize{SeedEdit} & 8.45 & 8.63 & 8.24 & 8.37 & 8.42 & 8.36 & 8.41 \\ \midrule
      \rowcolor{lightgray}\multicolumn{9}{c}{\textit{\normalsize{Multi-Agent Framework}}} \\
     \multirow{3}{*}{\textbf{\normalsize{\shortstack{FLUX.1\\-Kontext-dev}}}} & \normalsize{Original} & 6.29 & 6.49 & 6.91 & 7.33 & 7.95 & 8.31 & 7.21 \\
      &  \normalsize{{ImageEdit-R1} (w/ RL)} & 7.18 & 7.31 & 6.97 & 7.12 & 7.26 & 7.71 & 7.26$^{\color{+}+0.05}$ \\
        &  \cellcolor{lightblue}\normalsize{\texttt{ImageEdit-R1}} & \cellcolor{lightblue}7.92 &  \cellcolor{lightblue}7.63 &  \cellcolor{lightblue}8.02 &  \cellcolor{lightblue}8.24  &  \cellcolor{lightblue}\underline{8.63} &  \cellcolor{lightblue}\textbf{8.94} &  \cellcolor{lightblue}8.23$^{\color{+}+1.02}$  \\\midrule
     \multirow{3}{*}{\textbf{\normalsize{\shortstack{Qwen\\-Image-Edit}}}} & \normalsize{Original} & 8.12 & 8.27 & 8.33 & 8.46 & 8.47 & 8.62 & 8.39 \\
      &  \normalsize{{ImageEdit-R1} (w/ RL)} & 7.23 & 7.84 & 8.17 & 7.93 & 7.41 & 7.38 & 7.66$^{\color{-}-0.73}$ \\
        &  \cellcolor{lightblue}\normalsize{\texttt{ImageEdit-R1}} & \cellcolor{lightblue}8.75 &  \cellcolor{lightblue}\underline{8.91} &  \cellcolor{lightblue}\textbf{8.78} & \cellcolor{lightblue}\textbf{8.84}  &  \cellcolor{lightblue}\textbf{8.86} &  \cellcolor{lightblue}\underline{8.93} &  \cellcolor{lightblue}\textbf{8.85}$^{\color{+}+0.46}$ \\\midrule
        \multirow{3}{*}{\textbf{\normalsize{\shortstack{Nano\\Banana}}}} & \normalsize{Original} & 8.66 & 8.80 & 7.93 & 8.29 & 7.79 & 8.47 & 8.32 \\
      &  \normalsize{{ImageEdit-R1} (w/ RL)} & 8.49 & 8.61 & 7.89 & 7.91 & 8.03 & 7.88 & 8.14$^{\color{-}-0.18}$ \\
        &  \cellcolor{lightblue}\normalsize{\texttt{ImageEdit-R1}} & \cellcolor{lightblue}\underline{8.94} &  \cellcolor{lightblue}\textbf{8.92} &  \cellcolor{lightblue}\underline{8.57} &  \cellcolor{lightblue}\underline{8.58}  &  \cellcolor{lightblue}8.44 &  \cellcolor{lightblue}8.51 &  \cellcolor{lightblue}\underline{8.66}$^{\color{+}+0.34}$ \\
    \bottomrule  
  \end{tabular}}
  \label{tab:main}
\end{table*}

\begin{table}[hbpt]
  \centering
  \caption{Evaluation prompts when using LLM-as-a-Judge.}
  \footnotesize
  \renewcommand{\arraystretch}{0.85} % tighter line spacing
  \begin{tabular}{p{13.3cm}}
    \toprule
    You are an expert image quality evaluator. You will be shown two images: \\
    1. The ORIGINAL image (first image) \\
    2. The EDITED image (second image) \\

    \textbf{Edit Request:} "\{edit\_request\}" \\

    Please evaluate how well the edited image fulfills the edit request compared to the original image. \\

    \textbf{Evaluation Criteria:} \\
    1. \textbf{Request Fulfillment} (0--4 points): Did the edit successfully accomplish what was requested? \\
    \quad - 4: Perfectly fulfills the request \\
    \quad - 3: Mostly fulfills the request with minor issues \\
    \quad - 2: Partially fulfills the request \\
    \quad - 1: Barely fulfills the request \\
    \quad - 0: Does not fulfill the request \\

    2. \textbf{Image Quality} (0--3 points): Is the edited image realistic, coherent, and high quality? \\
    \quad - 3: Excellent quality, looks natural and professional \\
    \quad - 2: Good quality with minor artifacts or inconsistencies \\
    \quad - 1: Noticeable quality issues or artifacts \\
    \quad - 0: Poor quality, obviously edited \\

    3. \textbf{Preservation} (0--3 points): Are unrelated parts of the image properly preserved? \\
    \quad - 3: All unrelated areas perfectly preserved \\
    \quad - 2: Minor changes to unrelated areas \\
    \quad - 1: Significant unwanted changes \\
    \quad - 0: Major damage to unrelated areas \\

    Be strict but fair in your evaluation. A score of 10 means perfect execution in all aspects. \\
    \bottomrule
  \end{tabular}
  \label{table:testset}
\vspace{-0.4cm}
\end{table}

\paragraph{Evaluation Metrics} Evaluating the quality of edited images poses significant challenges, as human evaluation is both costly and subject to variability. To address this, we adopt the LLM-as-a-judge paradigm for automated assessment of image editing quality. Specifically, we utilize GPT-4o~\citep{hurst2024gpt} and Gemini-2.5-Flash~\citep{comanici2025gemini} as the backbone models for evaluation. The prompt template used for this assessment is detailed in Table~\ref{table:testset}. The evaluation score ranges from 0 to 10, where a score of 10 indicates a perfect edit and 0 represents the poorest quality edit.

% If sample 10,000 examples from each dataset. For evaluation:

% Gemini-2.5-flash: \$300

% GPT-4o: \$2000

% O4-mini: \$900

\paragraph{Implement Details}
 For reinforcement learning training, we use the EasyR1~\citep{zheng2025easyr1} framework built on verl~\citep{sheng2024hybridflow}, with specialized support for VLMs. Experiments are conducted using eight 80GB NVIDIA A100 GPUs with a global batch size of $32$, a rollout temperature of $1.0$, a consistent learning rate of $1\times10^{-6}$, and 8 rollouts. 

\begin{figure*}[t]
    \centering
    \includegraphics[width=0.98\textwidth]{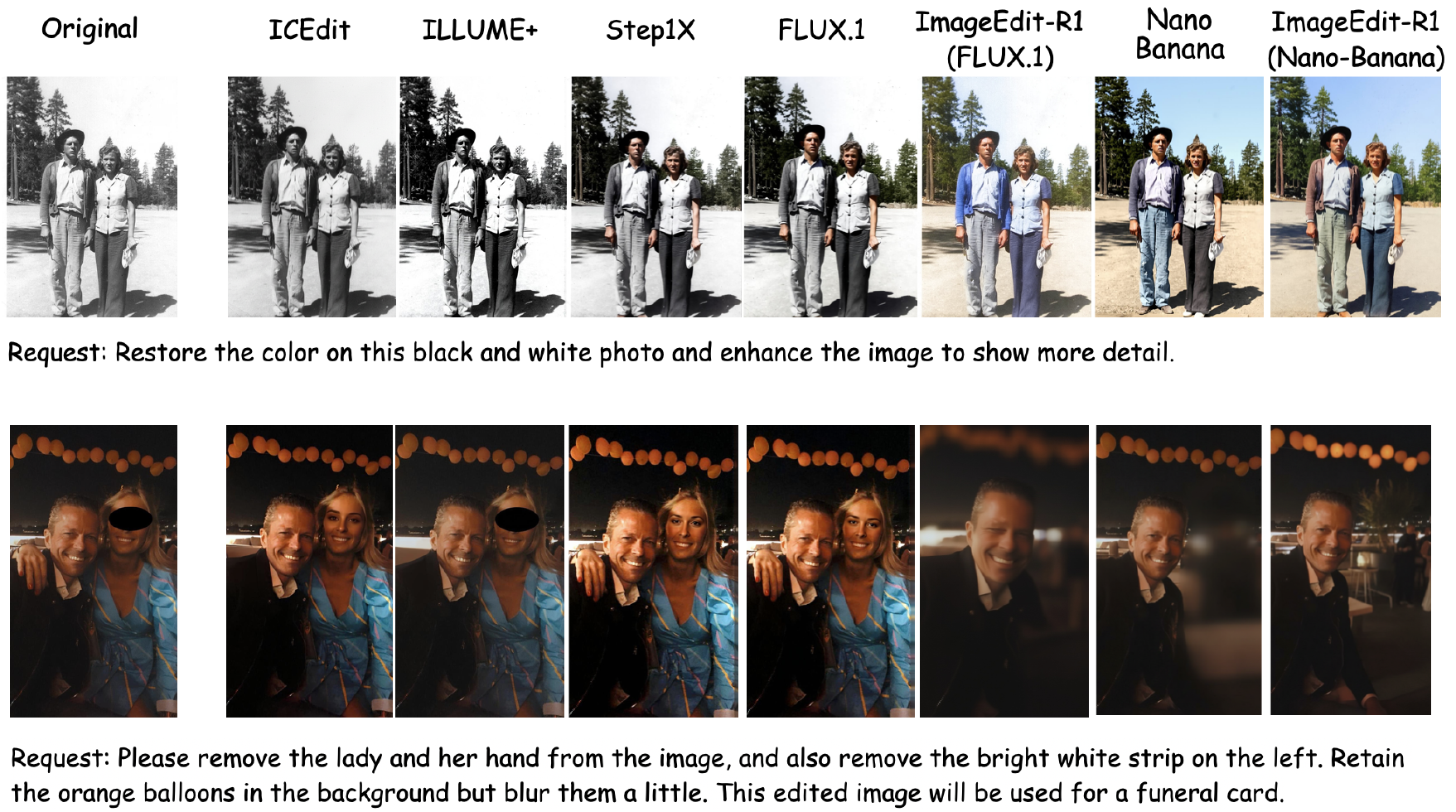}
    \includegraphics[width=0.98\textwidth]{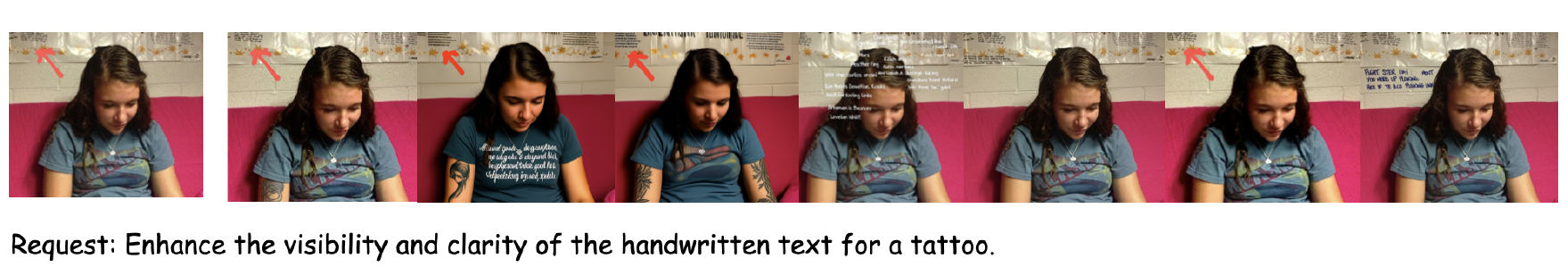}
    \caption{Representative examples demonstrating the performance of \texttt{ImageEdit-R1} compared to baselines on complex editing tasks.}
    \label{fig:example}
\end{figure*}

\subsection{Main Results}

\paragraph{\texttt{ImageEdit-R1} consistently improves the performance of base models.}
As shown in Table~\ref{tab:main}, integrating \texttt{ImageEdit-R1} leads to consistent performance gains across all base models and benchmarks. For example, it improves the average score of FLUX.1 from 7.21 to 8.23 (+1.02), Qwen-Image-Edit from 8.39 to 8.85 (+0.46), and NanoBanana from 8.32 to 8.66 (+0.34). These improvements are observed across all datasets and evaluator models, demonstrating that our decomposition framework enhances instruction alignment and editing quality regardless of the underlying editing backbone.

\paragraph{RL training is important for improving the performance of \texttt{ImageEdit-R1}.}
Table~\ref{tab:main} shows that the multi-agent framework alone (\texttt{ImageEdit-R1 (w/o RL)}) provides only marginal improvements or even results in performance drops compared to the original base models. For example, FLUX.1-Kontext-dev shows a slight increase from 7.21 to 7.26, while Qwen-Image-Edit and NanoBanana exhibit decreases in average performance. In contrast, when reinforcement learning is applied to train the decomposition agent, the performance improves significantly across all models. The average score of FLUX.1 increases to 8.23, Qwen-Image-Edit reaches 8.85, and NanoBanana achieves 8.66. These results highlight that reinforcement learning is essential for fully realizing the benefits of the multi-agent framework, as it enables more effective instruction decomposition and leads to higher editing quality.

\paragraph{\texttt{ImageEdit-R1} outperforms individual open-source and closed-source models.}  

As shown in Table \ref{tab:main}, \texttt{ImageEdit-R1} significantly outperforms both single DiT-based models and closed-source systems. While the best closed-source model (GPT-4o) achieves an average score of 8.47, \texttt{ImageEdit-R1} reaches 8.85 with Qwen-Image-Edit. In contrast, single-model baselines perform notably worse, with averages ranging from 6.33 to 7.04. This highlights the advantage of our multi-agent framework in achieving stronger instruction alignment and edit quality without relying on proprietary models.

\paragraph{Reward signals increase consistently and rapidly during training.}

\begin{figure*}[t]
  \centering
  \hspace{-0.2cm}
  \begin{minipage}{0.95\linewidth}
    \centering
    \includegraphics[width=\textwidth]{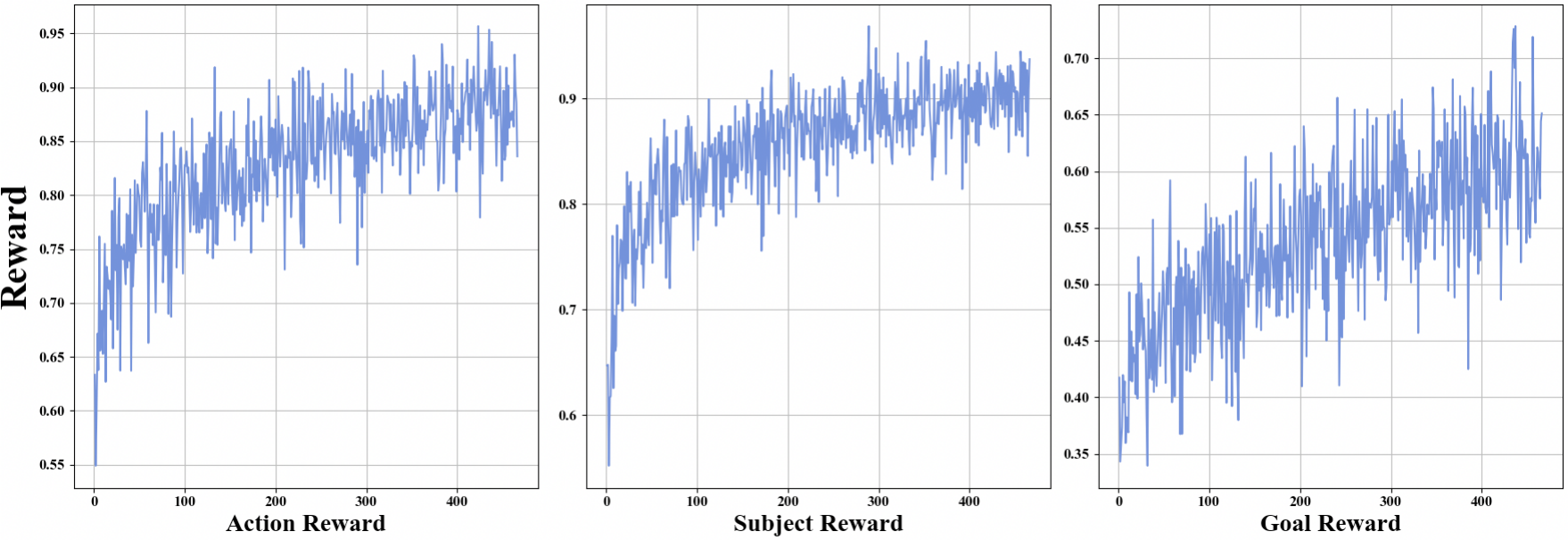}
    \caption{Training rewards of \textit{decomposition agent} on Qwen2.5-VL-7B-Instruction.}
    \label{fig:training}
  \end{minipage}%
  % right:tab
  
\vspace{-0.2cm}
\end{figure*}

Figure \ref{fig:training} illustrates the training progress of the sequencing agent, showing consistent improvements across action, subject, and goal rewards. The agent quickly learns to identify correct actions and subjects, with corresponding rewards stabilizing above 0.9. Goal reward improves more gradually, reflecting the higher complexity of capturing user intent, but still shows a clear upward trend. These results indicate that reinforcement learning with GRPO effectively enhances the agent’s ability to generate accurate and interpretable editing sequences.

%\begin{minipage}{0.3\linewidth}
%    \centering
%    \includegraphics[width=0.98\textwidth]{Figures/training_3b.pdf}
%    \caption{Training rewards of the Qwen2.5-VL-3B-Instruction variant.}
%    \label{fig:training_3b}
%  \end{minipage}%

% \begin{figure*}[t]
%     \centering
%     \includegraphics[width=0.98\textwidth]{Figures/training.pdf}
%     \caption{Training rewards of \texttt{ImageEdit-R1} \textit{decomposition agent} on Qwen2.5-VL-7B-Instruction.}
%     \label{fig:training}
% \end{figure*}

% \begin{figure}[t]
%     \centering
%     \includegraphics[width=0.46\textwidth]{Figures/training_3b.pdf}
%     \caption{Training progress of the Qwen2.5-VL-3B model variant, showing steady reward improvements but lower performance than the Qwen2.5-VL-7B model.}
%     \label{fig:training_3b}
% \end{figure}

\subsection{Concrete Example}

Figure~\ref{fig:example} presents qualitative comparisons highlighting the advantages of \textit{ImageEdit-R1} over existing baselines on complex multi-step editing tasks. In the first example, where the request is to restore color and enhance an old black and white photo, the \textit{ImageEdit-R1} variants produce more realistic tones, sharper textures, and improved visual clarity. Notably, the Nano Banana version recovers natural skin tones and environmental details more effectively than other methods. In contrast, baselines such as ICEdit and ILLUME+ tend to produce muted or inaccurate colors and fail to enhance image quality. The second example involves removing a specific person and a bright strip while preserving and slightly blurring the orange balloons in the background. \textit{ImageEdit-R1} successfully performs all parts of the instruction with minimal artifacts and strong spatial consistency. Other models often leave unwanted remnants or fail to blur the background correctly. 
% These results demonstrate that \textit{ImageEdit-R1} is capable of understanding nuanced instructions and generating visually coherent and instruction-aligned edits. 
In the third example, the instruction is to enhance the visibility and clarity of the handwritten text on the wall, intended as a tattoo reference. \textit{ImageEdit-R1} correctly identifies the target region and enhances the handwriting with improved sharpness and contrast, making it more legible while preserving the natural texture of the background. In contrast, baseline models often fail to localize the text accurately or introduce visual artifacts. These results demonstrate that \textit{ImageEdit-R1} is capable of understanding nuanced instructions and generating visually coherent and instruction-aligned edits.

\section{Analysis}

\subsection{Human Alignment}

% \begin{table}[hbpt]
%   \centering
% \renewcommand{\arraystretch}{1.3}
% \caption{Human vs. VLM-based evaluation. Top rows show average scores; bottom row shows their correlation.}
% \footnotesize
% \setlength{\tabcolsep}{3.3pt}
% \scalebox{0.95}{
%   \begin{tabular}{l|c|c|c}
%     \toprule
%    \textbf{\normalsize{Method}}  &  {\textbf{{ Fulfillment} (0-4)}}  & {\textbf{{Quality} (0-3)}} & {\textbf{{Preservation} (0-3)}}   \\
%     \midrule 
%      \normalsize{Human} & \normalsize{2.84} & \normalsize{2.38} &  \normalsize{2.73} \\
%        \normalsize{VLM-based} & \normalsize{{3.43}} & {\normalsize{2.41}} & \normalsize{2.26}   \\ \midrule
%     \rowcolor{lightblue}\normalsize{Correlation} & \normalsize{0.66} & \normalsize{0.43} & \normalsize{0.59}  \\
%     \bottomrule  
%   \end{tabular}}
%   \label{tab:human-vs-ai}
% \end{table}

\begin{wrapfigure}{r}{0.6\linewidth}
  \centering
  \renewcommand{\arraystretch}{1.3}
  \setlength{\tabcolsep}{3.3pt}
  \footnotesize
  \caption{Human vs. VLM-based evaluation. Top rows show average scores; bottom row shows their correlation.}
  \label{tab:human-vs-ai}
  \vspace{-0.25em}
  \scalebox{0.9}{
    \begin{tabular}{l|c|c|c}
      \toprule
      \textbf{\normalsize Method} &
      \textbf{Fulfillment (0--4)} &
      \textbf{Quality (0--3)} &
      \textbf{Preservation (0--3)} \\
      \midrule
      \normalsize Human & \normalsize 2.84 & \normalsize 2.38 & \normalsize 2.73 \\
      \normalsize VLM-based & \normalsize 3.43 & \normalsize 2.41 & \normalsize 2.26 \\
      \midrule
      \rowcolor{lightblue}\normalsize Correlation &
      \normalsize 0.66 & \normalsize 0.43 & \normalsize 0.59 \\
      \bottomrule
    \end{tabular}
  }
  \vspace{-0.35em}
\end{wrapfigure}

We randomly sample 100 editing instructions along with their corresponding original and edited images, and ask three human annotators to score each result using the same evaluation prompt described in Table~\ref{table:testset}. We then compare the average scores from human evaluations with those produced by the LLMs-as-a-judge setup and compute their correlation. As shown in Table~\ref{tab:human-vs-ai}, VLMs tend to assign slightly higher scores for fulfillment and quality, but lower scores for preservation. Despite these differences in score distribution, the correlation between human and model evaluations is relatively strong, particularly for fulfillment (0.66) and preservation (0.59). This indicates that VLM-based evaluation aligns reasonably well with human judgment, especially in assessing instruction completion and content preservation.

\subsection{Ablation Analysis}

\paragraph{Decomposition and sequencing models}

% \begin{table}[t]
%   \centering
% \renewcommand{\arraystretch}{1.3}
% \caption{Ablation analysis of the decomposition agent in \texttt{ImageEdit-R1}, comparing performance across a smaller model and models from different series. The backbone diffusion model is {FLUX.1-Kontext-dev}, and evaluation is conducted using {GPT-4o}.}
% \footnotesize
% \setlength{\tabcolsep}{3.3pt}
% \scalebox{0.98}{
%   \begin{tabular}{l|c|c|c|l}
%     \toprule
%    \textbf{\normalsize{Method}}  &  {\textbf{\normalsize{PSR}}}  & {\textbf{\normalsize{RealEdit}}} & {\textbf{\normalsize{UltraEdit}}}  & {\textbf{\normalsize{Avg.}}}  \\
%     \midrule 
%     \rowcolor{lightgray}\normalsize{Original} & 6.29 & 6.91 &  7.95 & 7.05 \\
%      \normalsize{Qwen2.5-VL-7B} & 7.92 & 8.02 &  \textbf{8.63} & 8.19 \\
%        %\normalsize{Qwen2.5-VL-3B} & 7.53 & 7.68 & 8.34 & 7.85  \\
%         \normalsize{Qwen3-VL-8B} & \textbf{7.94} & \textbf{8.33} & 8.61 & \textbf{8.29}  \\
%     \bottomrule  
%   \end{tabular}}
%   \label{tab:model_abla}
% \end{table}

\begin{table*}[t]
\centering

\begin{minipage}[t]{0.49\textwidth}
  \centering
  \renewcommand{\arraystretch}{1.3}
  \caption{Ablation analysis of the decomposition agent in \texttt{ImageEdit-R1}, comparing performance across a smaller model and models from different series. The backbone diffusion model is {FLUX.1-Kontext-dev}, and evaluation is conducted using {GPT-4o}.}
  \footnotesize
  \setlength{\tabcolsep}{3.3pt}
  \scalebox{0.85}{
  \begin{tabular}{l|c|c|c|l}
    \toprule
    \textbf{\normalsize{Method}}  &  {\textbf{\normalsize{PSR}}}  & {\textbf{\normalsize{RealEdit}}} & {\textbf{\normalsize{UltraEdit}}}  & {\textbf{\normalsize{Avg.}}}  \\
    \midrule
    \rowcolor{lightgray}\normalsize{Original} & 6.29 & 6.91 &  7.95 & 7.05 \\
    \normalsize{Qwen2.5-VL-7B} & 7.92 & 8.02 &  \textbf{8.63} & 8.19 \\
    \normalsize{Qwen3-VL-8B} & \textbf{7.94} & \textbf{8.33} & 8.61 & \textbf{8.29}  \\
    \bottomrule
  \end{tabular}}
  \label{tab:model_abla}
\end{minipage}\hfill
\begin{minipage}[t]{0.49\textwidth}
  \centering
  \renewcommand{\arraystretch}{1.3}
  \caption{Ablation analysis of the decomposition model with and without goals of editing in the reward.}
  \footnotesize
  \setlength{\tabcolsep}{3.3pt}
  \scalebox{0.85}{
  \begin{tabular}{l|c|c|c|l}
    \toprule
    \textbf{\normalsize{Method}}  &  {\textbf{\normalsize{PSR}}}  & {\textbf{\normalsize{RealEdit}}} & {\textbf{\normalsize{UltraEdit}}}  & {\textbf{\normalsize{Avg.}}}  \\
    \midrule
    \rowcolor{lightgray}\normalsize{Original} & 6.29 & 6.91 &  7.95 & 7.05 \\
    \normalsize{Qwen2.5-VL-7B} & 7.92 & 8.02 &  8.63 & 8.19 \\
    \normalsize{$\hookrightarrow$ without goal} & 7.62 & 7.71 & 8.43 & 7.92  \\
    \bottomrule
  \end{tabular}}
  \label{tab:goal_abla}
\end{minipage}

\end{table*}

We conduct an ablation analysis to investigate the impact of different VLMs on the performance of the decomposition agent in \texttt{ImageEdit-R1}. 
%Specifically, we evaluate a smaller model, Qwen2.5-3B-VL-Instruction, from the same series as our primary model. As shown in Figure~\ref{fig:training_3b}, the 3B model exhibits steady improvements across all reward dimensions throughout training. The action reward increases rapidly during the early stages and stabilizes around 0.7, reflecting the model’s ability to effectively identify and execute core editing actions. The subject reward improves more gradually, indicating progressive gains in object grounding. The goal reward, which captures alignment with user intent, remains the most challenging to optimize and shows slower growth, reflecting the difficulty of learning high-level semantics with a smaller model. While these results confirm that the 3B model can learn meaningful decomposition behavior, its performance remains limited compared to the larger 7B model, particularly in abstract and goal-oriented tasks.

Table~\ref{tab:model_abla} summarizes the quantitative results of the ablation study across three benchmark datasets: PSR, RealEdit, and UltraEdit. The original agent, which operates without decomposition, achieves an average score of 7.05. Incorporating decomposition with the Qwen2.5-VL-7B backbone yields substantial improvements, reaching an average performance of 8.19. 
%When using the smaller Qwen2.5-VL-3B model, performance slightly declines to 7.85, though it still outperforms the original agent by a notable margin. These results suggest that decomposition consistently enhances generation quality, but larger models are better equipped to exploit its benefits. 
The Qwen3-VL-8B model achieves the highest average score (8.29), slightly surpassing Qwen2.5-VL-7B. However, this improvement is marginal, as the Qwen2.5-VL-7B model already approaches the performance upper bound due to effective reinforcement learning, as shown in Figure~\ref{fig:training}. This suggests that once the agent is well-trained, scaling the VLM offers limited additional gains. Overall, the results highlight the central role of reinforcement learning in driving performance, with model size offering diminishing returns beyond a certain point.

\paragraph{Multi-turn or single turn}

% \begin{figure}[t]
%     \centering
%     \includegraphics[width=0.6\textwidth]{Figures/turn.pdf}
%     \caption{Comparison of multi-turn and single-turn strategies in the image editing process.}
%     \label{fig:multi_single_turn}
% \vspace{-0.2cm}
% \end{figure}

\begin{wrapfigure}{r}{0.6\textwidth}
  \centering
  \vspace{-0.6\baselineskip} % adjust if needed
  \includegraphics[width=0.60\textwidth]{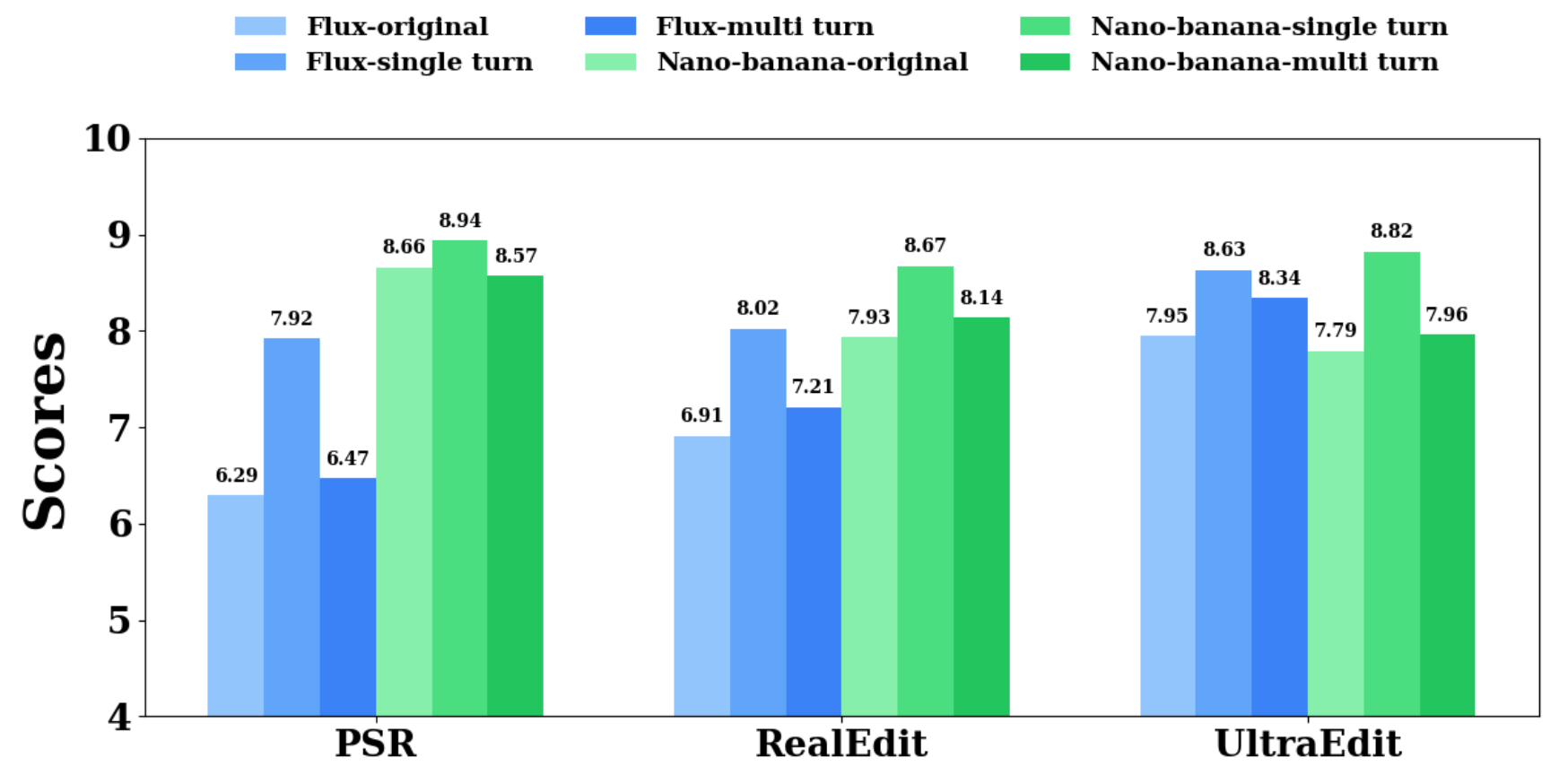}
  \caption{Comparison of multi-turn and single-turn strategies in the image editing process.}
  \label{fig:multi_single_turn}
  \vspace{-0.2cm}
\end{wrapfigure}

As the sequencing agent generates multiple sub-requests, there are two strategies for forwarding them to the image editing model: multi-turn, where the requests are sent sequentially in multiple steps, and single-turn, where all sub-requests are sent together in a single input. As shown in Figure~\ref{fig:multi_single_turn}, the single-turn approach consistently outperforms the multi-turn strategy across all benchmarks (PSR, RealEdit, and UltraEdit) and both backbone models (Flux and Nano-Banana). For example, under the Flux configuration, the single-turn method achieves a PSR score of 7.92 compared to only 6.47 for the multi-turn variant. A similar trend is observed with the Nano-Banana model, where single-turn execution leads to higher scores across all metrics. These results suggest that the multi-turn strategy suffers from compounding errors across editing rounds, as each sub-request is applied without global context or awareness of previous modifications. Moreover, sub-requests in the multi-turn setting may not be well-aligned with the visual state resulting from prior steps, leading to degraded edit quality. In contrast, the single-turn strategy benefits from executing all sub-requests in a unified context, allowing the editing model to better reason about dependencies and produce more coherent and effective image edits. This analysis highlights the advantage of holistic, single-step reasoning in the image editing pipeline.

\paragraph{With and without goals}

% \begin{figure*}[t]
%   \centering

%   \begin{subfigure}[t]{0.49\textwidth}
%     \centering
%     \includegraphics[width=0.95\linewidth]{Figures/without_goal.pdf}
%     \caption{Training progress of the decomposition model with and without goals of editing in the reward.}
%     \label{fig:training_goal}
%   \end{subfigure}\hfill
%   \begin{subfigure}[t]{0.49\textwidth}
%     \centering
%     \includegraphics[width=0.95\linewidth]{Figures/training_data.pdf}
%     \caption{Ablation analysis of training data for the decomposition agents in the RL training data.}
%     \label{fig:training_data}
%   \end{subfigure}

%   % \vspace{-0.2cm} % if you still need it, put it here (outside subfigures)
% \end{figure*}

\begin{figure*}[t]
  \centering

  \begin{minipage}[t]{0.49\textwidth}
    \centering
    \includegraphics[width=0.95\linewidth]{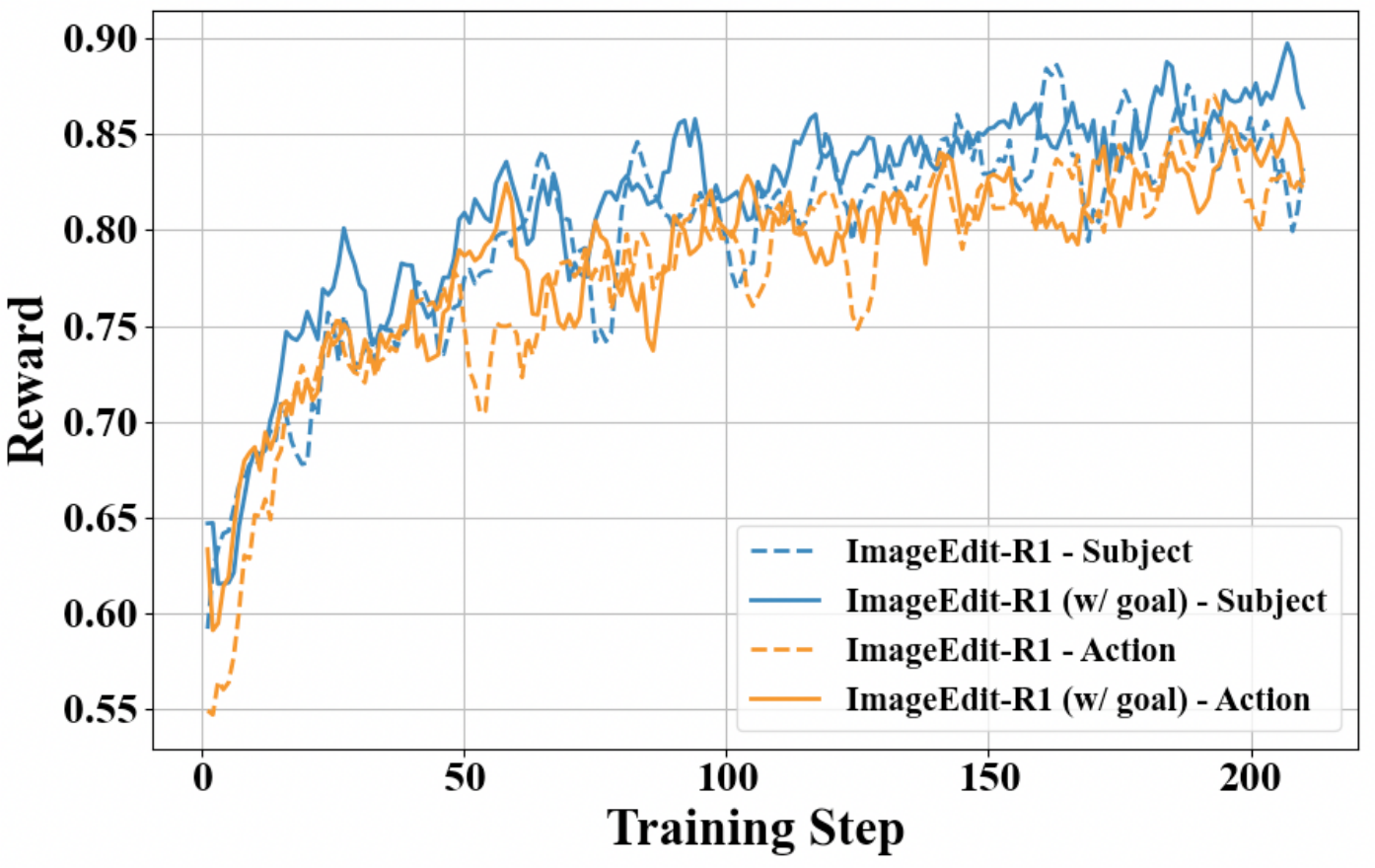}
    \caption{Training progress of the decomposition model with and without goals of editing in the reward.}
    \label{fig:training_goal}
  \end{minipage}\hfill
  \begin{minipage}[t]{0.49\textwidth}
    \centering
    \includegraphics[width=0.95\linewidth]{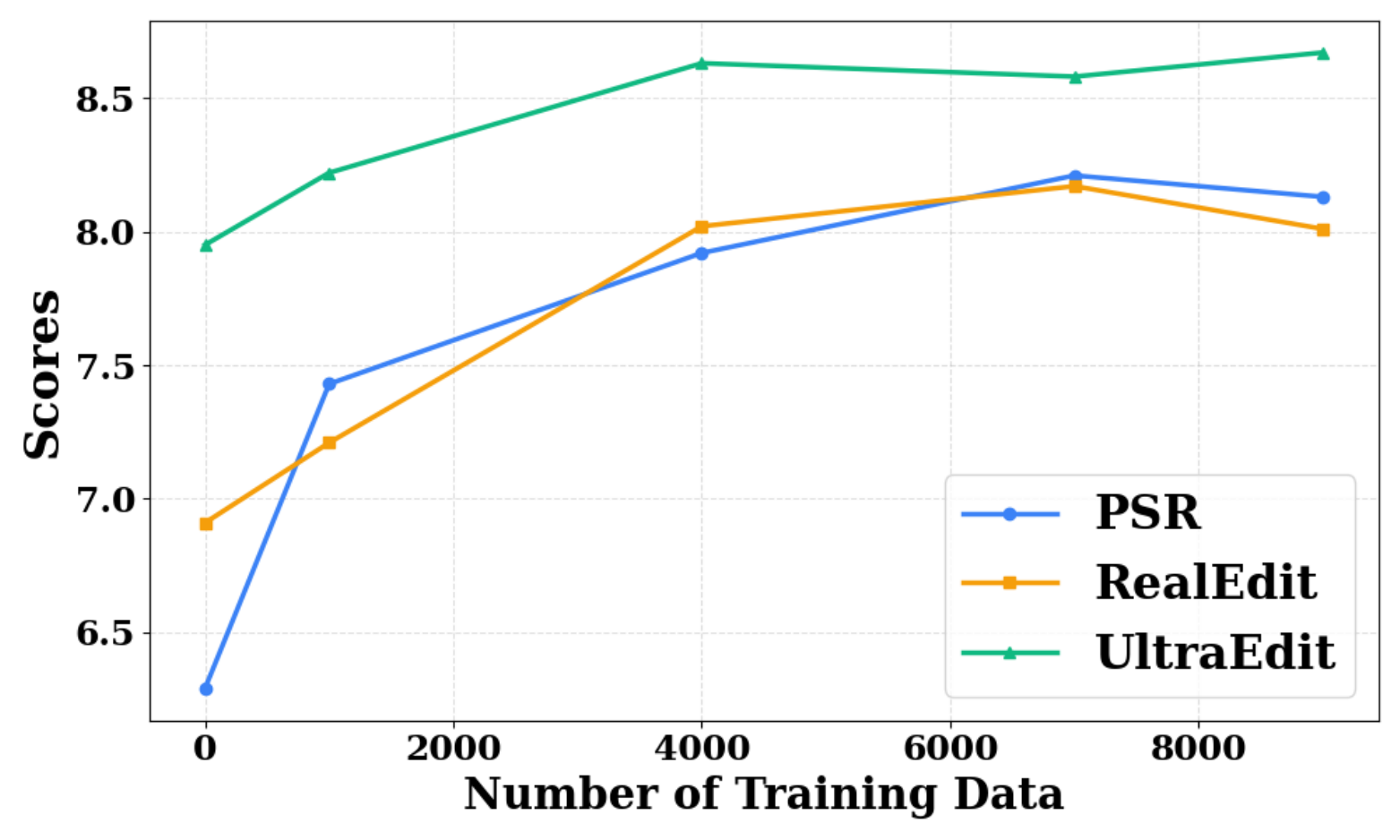}
    \caption{Ablation analysis of training data for the decomposition agents in the RL training data.}
    \label{fig:training_data}
  \end{minipage}

  % \vspace{-0.2cm} % optional, if you still need it
\end{figure*}

% \begin{figure}[t]
%     \centering
%     \includegraphics[width=0.46\textwidth]{Figures/without_goal.pdf}
%     \caption{Training progress of the decomposition model with and without goals of editing in the reward.}
%     \label{fig:training_goal}
% \end{figure}

% \begin{table}[t]
%   \centering
% \renewcommand{\arraystretch}{1.3}
% \caption{Ablation analysis of the decomposition model with and without goals of editing in the reward.}
% \footnotesize
% \setlength{\tabcolsep}{3.3pt}
% \scalebox{0.98}{
%   \begin{tabular}{l|c|c|c|l}
%     \toprule
%    \textbf{\normalsize{Method}}  &  {\textbf{\normalsize{PSR}}}  & {\textbf{\normalsize{RealEdit}}} & {\textbf{\normalsize{UltraEdit}}}  & {\textbf{\normalsize{Avg.}}}  \\
%     \midrule 
%     \rowcolor{lightgray}\normalsize{Original} & 6.29 & 6.91 &  7.95 & 7.05 \\
%      \normalsize{Qwen2.5-VL-7B} & 7.92 & 8.02 &  8.63 & 8.19 \\
%      \normalsize{$\hookrightarrow$ without goal} & 7.62 & 7.71 & 8.43 & 7.92  \\
%     \bottomrule  
%   \end{tabular}}
%   \label{tab:goal_abla}
% \vspace{-0.2cm}
% \end{table}

We perform an ablation study to investigate the impact of incorporating explicit goal supervision into the reward function of the decomposition agent. As shown in Figure~\ref{fig:training_goal}, both the goal-aware and goal-agnostic variants exhibit similar trends in subject and action rewards throughout training. This suggests that, during the learning process, the inclusion of goal information does not drastically affect the agent's ability to learn low-level sub-tasks such as identifying relevant subjects or performing action localization. However, despite their similar reward curves during training, the final image editing performance differs notably between the two variants. As reported in Table~\ref{tab:goal_abla}, the model trained with goal supervision achieves a higher average score of 8.19 across PSR, RealEdit, and UltraEdit benchmarks, compared to 7.92 for the model trained without goal-specific rewards. This gap highlights that goal conditioning plays a critical role in aligning the decomposition outputs with the overarching user intent. While both models can learn syntactically and visually grounded sub-requests, the goal-aware variant is better equipped to produce edits that are semantically consistent and contextually appropriate.

% \paragraph{Sequencing agents}

\paragraph{Training data}

% \begin{figure}[t]
%     \centering
%     \includegraphics[width=0.46\textwidth]{Figures/training_data.pdf}
%     \caption{Ablation analysis of training data for the decomposition agents in the RL training data.}
%     \label{fig:training_data}
% \vspace{-0.2cm}
% \end{figure}

Figure~\ref{fig:training_data} presents an ablation analysis of the impact of training data size for the decomposition agent during reinforcement learning, evaluated over three epochs on the \texttt{ImageEdit-R1} benchmark. As the number of training examples increases, performance consistently improves across all three evaluation metrics—PSR, RealEdit, and UltraEdit—indicating that larger datasets facilitate more effective learning of decomposition strategies. The most substantial gains are observed between 0 and 4000 training samples, where all metrics rise sharply. However, beyond 4000 examples, performance gains begin to plateau and remain relatively stable, suggesting that the decomposition agent has acquired most of the essential capabilities needed to decompose instructions into actions, subjects, and goals. These results highlight the importance of training data scale in the early phases of learning, while also suggesting diminishing returns as the dataset continues to grow.

\section{Related Work}

\paragraph{Instruction-Based Image Editing}

Instruction-based image editing aims to guide generative models in modifying images according to user-provided instructions. Early methods such as InstructPix2Pix~\citep{brooks2023instructpix2pix} and InstructEdit~\citep{wang2023instructedit} train text-to-image models using paired instruction-image datasets, predominantly built upon diffusion models~\citep{ho2020denoising, song2020denoising, peebles2023scalable}. While effective for simple edits, these single-model approaches often struggle with complex, multi-step tasks and lack fine-grained regional control. Recent advances have increasingly integrated multimodal large language models~\citep{liu2023visual} into editing pipelines, leveraging their superior comprehension capabilities to interpret images and refine instructions for more precise edits. Notable works including Qwen2.5-VL~\citep{bai2025qwen25vltechnicalreport}, ILLUME+~\citep{huang2025illumeilluminatingunifiedmllm}, and Intern-VL~\citep{chen2024internvl} have significantly improved image quality and text-image alignment. General-purpose models such as Gemini~\citep{comanici2025gemini} and GPT-4o~\citep{hurst2024gpt} demonstrate strong vision-language consistency through large-scale training, reflecting a shift toward integrated editing frameworks from isolated systems.
High-quality instruction-image datasets have also emerged to support model training. Recent datasets such as PSR~\citep{taesiri2025understanding} and RealEdit~\citep{sushko2025realedit} focus on real-world scenarios, containing diverse editing cases ranging from simple object manipulations to complex contextual changes, and typically comprise thousands to hundreds of thousands of pairs with careful visual-textual alignment.

\paragraph{Reinforcement Learning in LLMs and VLMs}

Reinforcement learning has become essential for enhancing LLMs and vision-language models through alignment and reasoning optimization. Reinforcement Learning from Human Feedback (RLHF) aligns outputs with human preferences~\citep{ouyang2022training}, while RLVR leverages verifiable rewards to improve reasoning. These approaches transform models from passive generators into active decision-makers that optimize long-term outcomes.
In image editing, RL optimizes editing policies through reward-based training. Pioneering work such as A2-RL~\citep{li2018a2} formulates editing as sequential decision-making. Recent advances include methods for stable multi-objective optimization, such as RewardEdit~\citep{gu2024multi} and Instructrl4pix~\citep{li2024instructrl4pix}, which employ multi-dimensional reward models for automated feedback, and GRPO. These methods~\citep{black2023training} enable agents to learn editing operations that maximize predefined reward functions while maintaining output quality.

\section{Conclusion}

We introduce \texttt{ImageEdit-R1}, a modular, multi-agent framework for instruction-based image editing that formulates the task as a sequential decision-making process. By using RL to train a decomposition agent and coordinating with sequencing and editing agents, our system translates natural language instructions into interpretable editing steps. 
\texttt{ImageEdit-R1} improves alignment with user intent, visual quality, and flexibility across diverse editing backbones, all without modifying the underlying models. 
Through extensive experiments and ablations, we demonstrate the benefits of structured reasoning, goal conditioning, and agent collaboration. 

% \section*{Limitation}

% Training of the decomposition agent relies on decompositions from existing datasets, which may not accurately reflect true user intent or optimal editing strategies. These predefined decompositions can introduce bias and limit the agent’s ability to learn more flexible or user-aligned behaviors. Furthermore, the editing tools used in the framework are relatively coarse, which constrains the precision and expressiveness of the edits. Integrating more fine-grained tools, such as those available in Photoshop, could significantly enhance the system’s capability and real-world applicability.

\bibliography{colm2026_conference}
\bibliographystyle{colm2026_conference}

% \appendix
% \section{Appendix}
% You may include other additional sections here.

\end{document}